\documentclass[10pt,twocolumn,letterpaper]{article}

\usepackage{cvpr}              % To produce the CAMERA-READY version
\usepackage[ruled,vlined]{algorithm2e}
\usepackage{multirow}
\usepackage{graphicx}  
\usepackage{array}
\usepackage{pifont}
\usepackage{amsmath}
\usepackage[table]{xcolor}

\definecolor{cvprblue}{rgb}{0.21,0.49,0.74}
\usepackage[pagebackref,breaklinks,colorlinks,allcolors=cvprblue]{hyperref}

\def\paperID{} % *** Enter the Paper ID here
\def\confName{CVPR}
\def\confYear{2026}

\title{BIT: Matching-based Bi-directional Interaction Transformation Network for Visible-Infrared Person Re-Identification}

\author{Haoxuan Xu \qquad Guanglin Niu\thanks{Corresponding author.} \\
School of Artificial Intelligence, Beihang University\\
{\tt\small xhaoxuan@buaa.edu.cn \qquad beihangngl@buaa.edu.cn}}
\newcommand\blfootnote[1]{%
  \begingroup
  \renewcommand\thefootnote{}%
  \footnotetext{#1}%
  \endgroup
}
\begin{document}
\maketitle
\blfootnote{\url{https://github.com/Xuan266/BIT}}
\begin{abstract}
Visible-Infrared Person Re-Identification (VI-ReID) is a challenging retrieval task due to the substantial modality gap between visible and infrared images. While existing methods attempt to bridge this gap by learning modality-invariant features within a shared embedding space, they often overlook the complex and implicit correlations between modalities. This limitation becomes more severe under distribution shifts, where infrared samples are often far fewer than visible ones. To address these challenges, we propose a novel network termed Bi-directional Interaction Transformation (BIT). Instead of relying on rigid feature alignment, BIT adopts a matching-based strategy that explicitly models the interaction between visible and infrared image pairs. Specifically, BIT employs an encoder-decoder architecture where the encoder extracts preliminary feature representations, and the decoder performs bi-directional feature integration and query aware scoring to enhance cross-modality correspondence. To our best knowledge, BIT is the first to introduce such pairwise matching-driven interaction in VI-ReID. Extensive experiments on several benchmarks demonstrate that our BIT achieves state-of-the-art performance, highlighting its effectiveness in the VI-ReID task.
\end{abstract}    
\begin{figure}[t]
  \centering
  
   \includegraphics[width=1.0\linewidth]{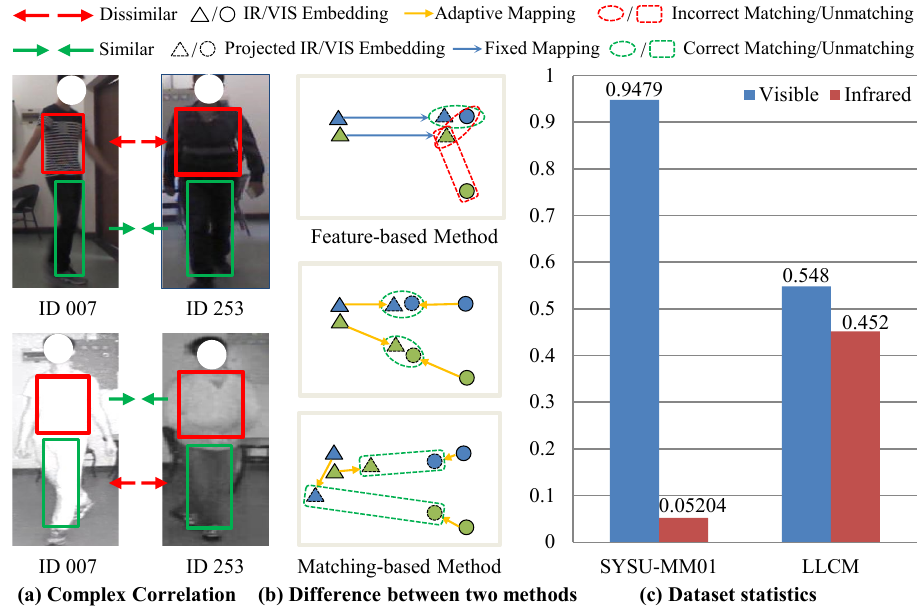}

   \caption{Motivation of Our Work. (a) Illustrates the cross-modality complexity in VI-ReID, where similar appearances in the visible domain may diverge significantly in infrared, and vice versa. (b) Highlights that fixed mappings in feature-based methods can cause incorrect matching, while our matching-based approach learns pair-specific adaptive transformations. (c) Reveals the modality imbalance in VI-ReID datasets, with far fewer infrared samples, further challenging feature-based learning and motivating our design.}
   \label{Fig1}
\end{figure}

\section{Introduction}
Person re-identification \cite{c2,c1} (Re-ID) aims to re-identify a pedestrian across different non-overlapping cameras. Most existing Re-ID methods focus on matching images captured by visible cameras. Nevertheless, conventional visible cameras struggle in poor illumination conditions (\textit{e.g.}, at night), which can introduce potential safety hazards. To address this issue, modern surveillance systems automatically switch to infrared mode to capture pedestrian information in low-light environments. Consequently, Visible-Infrared Person Re-ID (VI-ReID), which aims to match pedestrian images across different spectral domains, has gained increasing attention.  

Compared to conventional Re-ID, VI-ReID presents additional challenges due to the substantial modality discrepancy between visible and infrared images. This modality gap further exacerbates the effects of external factors such as viewpoint variations and background clutter, making reliable matching even more difficult. To mitigate these challenges, existing methods can be broadly categorized into image-level and feature-level approaches. Image-level methods attempt to bridge the gap by generating cross-modality images via style transformation, often using generative-based models \cite{c16, c17, c18, c19}. However, these approaches frequently introduce noise and suffer from increased computational costs. Alternatively, feature-level methods focus on learning modality-invariant representations by aligning visible and infrared features within a shared embedding space \cite{c14, c15, c16, c17}. While these methods enhance cross-modality feature consistency to some extent, they often depend on static alignment strategies and fail to generalize well under distribution shifts.  

Although existing approaches mitigate the modality gap to some extent, they often overlook the complex correlation between visible and infrared images. Infrared intensities reflect electromagnetic radiation just beyond the visible spectrum, influenced by factors such as color, material properties, and heat distribution, many of which go beyond the perceptual and modeling capabilities of current vision systems. As illustrated in Fig.~\ref{Fig1}(a), clothing with similar visible appearances may appear dramatically different in infrared, while those with distinct visible colors can appear nearly identical. This inconsistency challenges the reliability of static feature transformation methods, which are prone to overfitting. As shown in Fig.~\ref{Fig1}(b), existing feature-based methods typically learn a fixed mapping that directly aligns visible features with their corresponding infrared features. However, when two samples from different identities exhibit highly similar appearances in the infrared modality, fixed mappings tend to project both of them close to the same visible feature. In this case, the visually similar negative sample can also be drawn close, causing misidentification. In contrast, matching-based method offer greater flexibility by learning adaptive transformation patterns tailored to each visible-infrared pair. Rather than enforcing global alignment, these methods directly capture instance-specific correspondences, enabling more reliable and fine-grained matching across modalities.

% This inconsistency challenges the reliability of static feature transformation methods, which are prone to overfitting, particularly under distribution shifts.

More critically, VI-ReID datasets are often highly imbalanced, with significantly fewer infrared samples than visible ones, as shown in Fig.~\ref{Fig1}(c). Such imbalance severely limits the learning capacity of traditional feature-based models, which rely on dense, balanced training data to form robust modality-invariant embeddings. In contrast, pairwise matching paradigms naturally focus on relational modeling rather than global representation learning, making them more robust to training data disparity. These observations motivate us to explore a new matching-based method for VI-ReID that can directly exploit fine-grained correspondences across modalities, rather than mapping them into a shared space.

To this end, we propose a novel matching-based network, \underline{\textbf{B}}i-directional \underline{\textbf{I}}nteraction \underline{\textbf{T}}ransformation (BIT), which adaptively models cross-modality associations rather than rigid alignment. Specifically, BIT contains two core designs: (1) a \underline{\textbf{B}}i-directional \underline{\textbf{C}}ross \underline{\textbf{I}}nteraction (BCI) module within a dual-stream network that enables visible-infrared image pairs to extract complementary information and dynamically reduce the modality gap; and (2) a \underline{\textbf{Q}}uery \underline{\textbf{A}}ware \underline{\textbf{Scoring}} (QA Scoring) module, which adaptively selects and aggregates the most relevant patch-level similarities for robust final matching. By leveraging these adaptive mechanisms, BIT not only mitigates the risk of overfitting from imbalanced training data, but also captures reliable local correspondences for fine-grained matching.

Our main contributions can be summarized as follows:
\begin{itemize}
    \item To our best knowledge, we are the first to propose a matching-based network BIT for VI-ReID that directly models cross-modality correspondence, adaptively mitigating the modality gap under data imbalance.
    \item We introduce a BCI module that enables effective fine-grained feature interaction across modalities.
    \item We propose a QA Scoring module that dynamically identifies reliable patch-level correspondences to enhance matching robustness.
    \item Extensive experiments on three popular VI-ReID benchmarks, including SYSU-MM01 \cite{c14}, LLCM \cite{c15} and RegDB \cite{c49}, demonstrate that BIT achieves state-of-the-art performance.
\end{itemize}

\section{Related Work}
\textbf{Single-Modality Person Re-ID.}  
Single-modality person Re-ID \cite{c1,c2,c3,c84,c85,c86,c87,c88} aims to match pedestrian images captured by different visible cameras. Although these approaches have achieved remarkable success on standard benchmarks, the increasing deployment of infrared cameras necessitates cross-modality matching. Some single-modality Re-ID approaches \cite{c5,c8,c9,c10,c11,c12,c13}, similar to ours, adopt a matching-based architecture. However, these approaches typically assume feature homogeneity and compute the final similarity score through a simple mapping, without performing the essential feature refinement needed to alleviate the modality discrepancy.

% For instance, Li \textit{et al.} \cite{c10} proposed a filter pairing neural network to address misalignment and occlusions, Ahmed \textit{et al.} \cite{c11} introduced a local neighborhood matching layer, and Suh \textit{et al.} \cite{c12} developed part-aligned bilinear representations to refine matching. Additionally, Shen \textit{et al.} \cite{c13} proposed a soft-alignment Kronecker-product matching module.

\noindent\textbf{Visible-Infrared Person Re-ID.}  
VI Re-ID \cite{c14, c15, c49,c76,c77, c51,c52,c53,c89} focuses on retrieving pedestrian images captured in both visible and infrared spectra. Existing methods can be broadly categorized into image-level and feature-level approaches.
Image-level approaches attempt to reduce modality discrepancies by transforming one modality into another, often using generative models \cite{c16, c17, c18, c19,c20,c21,c22,c74} that perform identity-preserving style transformation. These methods facilitate cross-modal alignment while compensating for the limited availability of paired training data. However, these approaches often suffer from high computational cost and inevitably introduce noises. 
Feature-level approaches \cite{c30,c46,c78} aim to learn modality-invariant \cite{c24,c25,c26,c27,c82,c29} or modality-compensated \cite{c81,c28,c36,c75,c83} representations in a shared feature space. For example, MID \cite{c35} introduced a novel modality-adaptive mixup and invariant decomposition method. CAJ \cite{c45} enhances model robustness to color variations via channel-wise augmentation. PMT \cite{c47} adopts a progressive learning framework with grayscale images serving as an intermediate modality to ease the modality gap. Additional, Part-based methods\cite{c37,c32,c31} have been explored to mitigate intra-class variance by leveraging local semantic consistency across modalities. 
\begin{figure*}[h]
    \centering
    \includegraphics[width=1.0\linewidth]{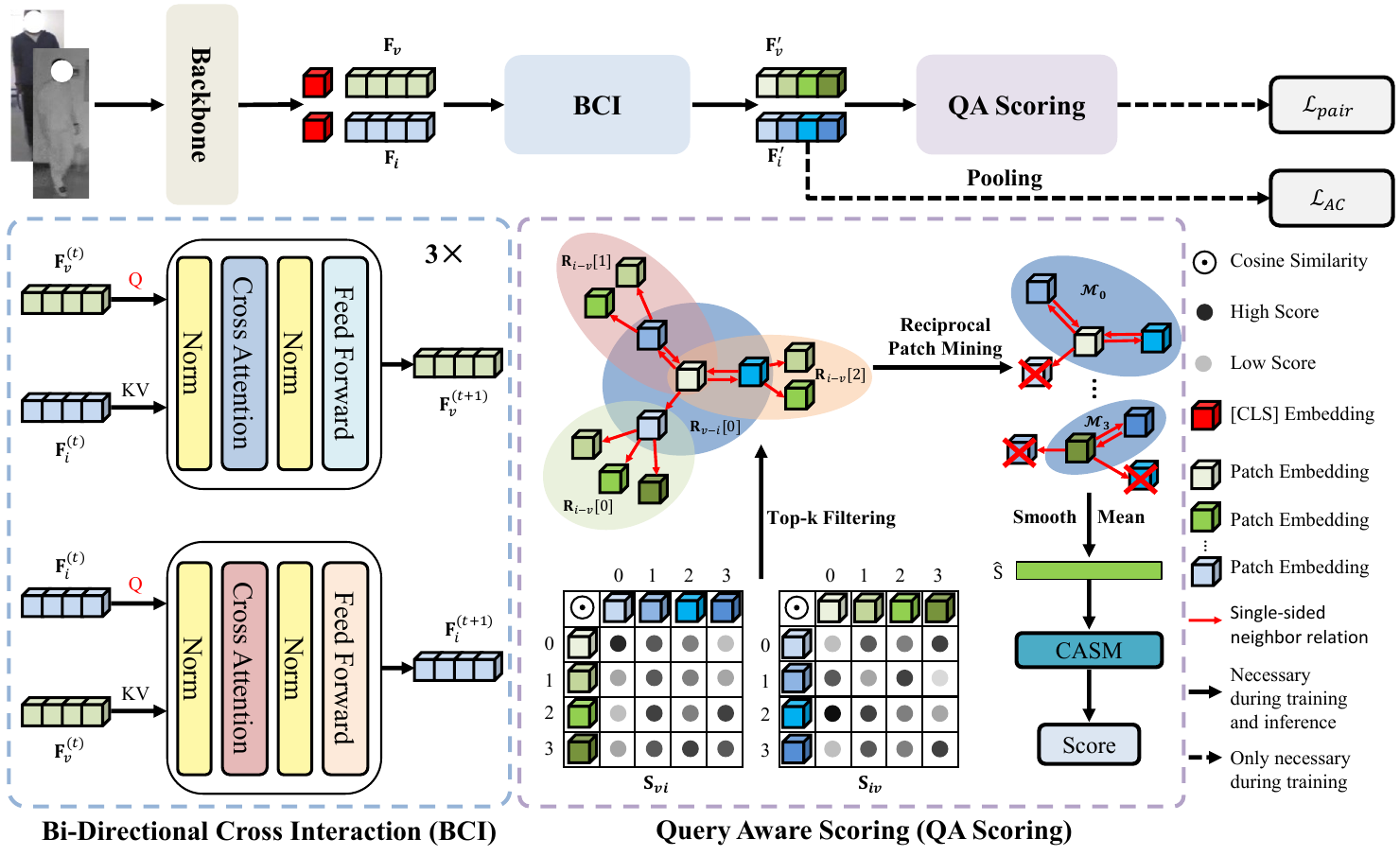}
\caption{Overall architecture of BIT. Given a visible-infrared image pair, we first extract features via a shared backbone. The Bi-Directional Cross Interaction (BCI) module enhances cross-modal representation through iterative feature exchange. Then, the Query Aware Scoring (QA Scoring) module computes fine-grained patch-wise similarities, adaptively identifying the most relevant response regions for final similarity score calculation.}
\label{fig2}
\end{figure*}
Despite advancements in both image-level and feature-level methods, most existing methods rely on predefined feature extraction pipelines or fixed pattern learning, which limits their ability to generalize across diverse environmental conditions. To address these limitations, we propose a matching-based network that adaptively aligns visible and infrared features, enabling robust and flexible cross-modality ReID.

% \section{Preliminaries}

% \subsection{Problem Formulation} 

% Let $\mathcal{Q} = \{\mathbf{q}_{i} \mid i=1,2, \dots, N_{q}\}$ denote the query set, where each $\mathbf{q}_i$ represents a pedestrian image captured in the infrared (or visible) spectrum. Similarly, let $\mathcal{G} = \{\mathbf{g}_{j} \mid j=1,2, \dots, N_{g}\}$ denote the gallery set, where each $\mathbf{g}_j$ corresponds to a pedestrian image acquired in the visible (or infrared) spectrum. Each image $x_i \in \mathcal{Q} \cup \mathcal{G}$ is associated with an identity label $y_j \in \mathcal{Y}$, where $\mathcal{Y}$ represents the set of unique person identities. The identity labels satisfy $y_i = y_j$ if and only if the query image $\mathbf{q}_i$ and the gallery image $\mathbf{g}_j$ belong to the same person.

% The objective of VI-ReID is to retrieve the correct identity from the gallery given a query image, despite the inherent modality discrepancy between infrared and visible images. Specifically, for a given query image $\mathbf{q}_i \in \mathcal{Q}$, the goal is to rank all images in $\mathcal{G}$ based on their similarity to $\mathbf{q}_i$, such that images of the same identity appear at the top of the ranked list.
\section{Method}
\subsection{Revisit Cross-Attention Mechanism} 
Cross-attention \cite{c23} is a fundamental mechanism widely utilized in vision and language tasks \cite{c80}. Unlike self-attention, which captures intra-modality relationships, cross-attention enables direct interaction between two distinct feature representations, making it particularly effective for cross-modality matching tasks.

% where $N$ represents the number of tokens (\textit{e.g.}, image patches) and $C$ denotes the embedding dimension, 

Formally, given a query embedding matrix $\mathbf{Q} \in \mathbb{R}^{N \times C}$ and a corresponding key-value embedding matrix $\mathbf{K} = \mathbf{V} \in \mathbb{R}^{N \times C}$, the cross-attention mechanism is defined as:
\begin{equation}
    \texttt{CrossAtt}(\mathbf{Q}, \mathbf{K}\mathbf{V}) = \texttt{s} \left( \frac{\mathbf{Q} \mathbf{W}_q (\mathbf{K} \mathbf{W}_k)^{\top}}{\sqrt{C}} \right) \mathbf{V} \mathbf{W}_v,
\end{equation}
where $\texttt{s}(\cdot)$ denotes softmax function, $\mathbf{W}_q, \mathbf{W}_k, \mathbf{W}_v \in \mathbb{R}^{C \times C}$ are learnable projection matrices that transform the query, key, and value embeddings, respectively. 

% By dynamically attending to the most relevant features from the paired modality, cross-attention effectively aligns information between different domains, reducing modality discrepancies and enhancing feature fusion for robust cross-modality representation learning.

\subsection{Overview}
The proposed BIT network is designed to tackle the core challenges of VI-ReID, which arise primarily from the substantial modality discrepancy between RGB and infrared images. As illustrated in Fig.~\ref{fig2}, BIT integrates three key components: (1) a shared backbone network for preliminary feature extraction, (2) a BCI module for fine-grained cross-modality feature fusion, and (3) a QA Scoring module for query aware similarity calculation. Notably, the backbone serves as the encoder, while the BCI and QA Scoring module jointly function as the decoder.

\subsection{Bi-Directional Cross Interaction}
To address the modality gap inherent in VI ReID, we introduce the BCI module, which adaptively models cross-modality associations through a cross-attention mechanism followed by iterative refinement. Instead of enforcing rigid feature alignment in a shared space, BCI enables visible and infrared representations to adaptively exchange and integrate complementary information. This design allows the model to preserve modality-specific characteristics while enhancing identity-discriminative cues across domains.

\noindent \textbf{Pair Construction within Batch.} To enable pairwise cross-modality interaction, we firstly construct all visible-infrared pairs within a batch. Given patch embeddings $\mathbf{F}_{v} \in \mathbb{R}^{B \times N \times C}$ for visible images and $\mathbf{F}_{i} \in \mathbb{R}^{B \times N \times C}$ for infrared images, we expand each visible sample by repeating it $B$ times along the batch dimension, resulting in $\mathbf{F}_{v}^{(0)} \in \mathbb{R}^{B^2\times N\times C}$, where each visible feature will be paired with every infrared feature in the batch. For the infrared features, we repeat the entire batch $B$ times to form $\mathbf{F}_{i}^{(0)} \in \mathbb{R}^{B^2\times N\times C}$. For clarity of illustration, Fig.~\ref{fig2} and the subsequent sections take a single image pair as an example.

The BCI module aims to establish adaptive correspondences between visible and infrared features. Specifically, we initiate interaction by computing mutually enhanced representations as follows:
\begin{equation} 
\begin{aligned}
\tilde{\mathbf{F}}_{v}^{(0)} &= \texttt{CrossAtt}(\mathbf{F}_{v}^{(0)}, \mathbf{F}_{i}^{(0)}), \\
\tilde{\mathbf{F}}_{i}^{(0)} &= \texttt{CrossAtt}(\mathbf{F}_{i}^{(0)}, \mathbf{F}_{v}^{(0)}),
\end{aligned} 
\end{equation} 
where each \texttt{CrossAtt}(Q, KV) module employs one modality as query and the other as key-value pair, enabling selective aggregation of complementary features. This design allows deeper mutual understanding across modalities.

\noindent\textbf{Bi-Directional Cross Interaction Block.}
To further refine the fused features, we introduce a Transformer-style \texttt{BCI Block}, which iteratively improves cross-modal representations in a multi-stage fashion. Each BCI block consists of two independent but interacting streams (visible and infrared), where each stream follows a residual structure with alternating cross-attention and feed-forward sub-layers. The two-stream architecture ensures that each modality retains its unique characteristics while being enriched through context-aware feature exchange.

Formally, the visible stream at the $t$-th stage is updated: \begin{equation}
    \hat{\mathbf{F}}_{v}^{(t)} = \mathbf{F}_{v}^{(t)} + \texttt{CrossAtt}(\text{LN}(\mathbf{F}_{v}^{(t)}), \text{LN}(\mathbf{F}_{i}^{(t)})),
\end{equation}
\begin{equation}
    \mathbf{F}_{v}^{(t+1)} = \hat{\mathbf{F}}_{v}^{(t)} + \texttt{MLP}(\text{LN}(\hat{\mathbf{F}}_{v}^{(t)})),
\end{equation}
where $\texttt{MLP}$ denotes a two-layer feed-forward network with GELU activation. The infrared stream is updated in the same manner by attending to visible features. 
% \begin{equation}
% \hat{\mathbf{F}}_{i}^{(t)} = \mathbf{F}_{i}^{(t)} + \texttt{CrossAtt}(\text{LN}(\mathbf{F}_{i}^{(t)}), \text{LN}(\mathbf{F}_{v}^{(t)}))
% \end{equation}
% \begin{equation}
% \mathbf{F}_{i}^{(t+1)} = \hat{\mathbf{F}}_{i}^{(t)} + \texttt{MLP}(\text{LN}(\hat{\mathbf{F}}_{i}^{(t)}))
% \end{equation}

Instead of relying on a single fusion, we stack $T$ BCI blocks to implement progressive refinement. At each stage, the visible and infrared representations evolve by absorbing semantic cues from each other while maintaining their individual domain priors. After $T$ iterations, the final features $\mathbf{F}_v^{'}$ and $\mathbf{F}_i^{'}$ serve as enhanced identity-discriminative representations for retrieval. This multi-stage refinement scheme empowers the model to gradually resolve cross-modality discrepancies and capture fine-grained correspondences.

% In essence, BCI transforms the problem of cross-modality matching from a domain-invariant learning task into a dynamic feature refinement process without aligning modalities by force, but through shared semantics and complementary structures.

\noindent\textbf{Global Interaction Constraint.}  
To guide the interaction and ensure that cross-modality aggregation is identity-consistent, we introduce an Aggregation Contrastive Loss. Let $i$ be the index of an arbitrary sample, the $\mathcal{L}_{AC}$ is:
\begin{equation}
\mathcal{L}_{AC} =  -\frac{1}{|P_i|} \sum_{p \in P_i} \log \frac{e^{(f_i \cdot f_p / \tau)}}{e^{(f_i \cdot f_p / \tau)}+\sum\limits_{j \in (N_i)} e^{(f_i \cdot f_j / \tau)}},
\label{eq8}
\end{equation}
where $f_i$ is the pooled representation from the final BCI block output $\mathbf{F}^{'}$, $P_i (N_i)$ denotes the set of samples sharing the same (different) identity label as sample $i$, and $\tau$ is a temperature parameter, which is fixed to $1/16$ according to \cite{c72}. $\mathcal{L}_{AC}$ acts as a regularization term that encourages positive pairs (from different modalities) to be drawn closer while pushing apart hard negatives.

The BCI module is specifically designed to tackle the core challenges of VI-ReID, where visible and infrared images often diverge drastically in terms of color, texture, and illumination. In contrast to previous methods that primarily emphasize global feature alignment, BCI introduces a bi-directional interaction paradigm based on localized cross-attention and iterative refinement. This mechanism promotes the learning of fine-grained, identity-discriminative correspondences by dynamically exchanging information between modalities. As a result, it not only suppresses reliance on superficial visual cues or background noise but also enhances generalization under modality discrepancies.

\subsection{Query Aware Scoring}
After obtaining refined visible and infrared features, we perform fine-grained matching through a novel QA Scoring module, which addresses a key limitation in traditional similarity computation that the equal treatment of all feature patches. In real-world VI-ReID scenarios, different query samples may rely on different visual cues depending on pose, occlusion, or background clutter. Thus, it is crucial to make the similarity estimation more query-specific and discriminative. The Algorithm.\ref{alg1} outlines the primary steps of the QA Scoring.

\begin{algorithm}[t]
\caption{Framework of QA Scoring}
\KwIn{Features $\mathbf{F}_v^{\prime}, \mathbf{F}_i^{\prime} \in \mathbb{R}^{N \times C}$, top-$k$ value $k$}
\KwOut{Final similarity score $\Psi$}

\textbf{Step 1:} Compute the patch-wise similarity matrices $\mathbf{S}_{vi}, \mathbf{S}_{iv}$ according to Eq.~(\ref{eq9}) \;

\textbf{Step 2:} Apply Top-$k$ filtering to construct the neighbors sub-graph for each patch using Eq.~(\ref{eq10}) \;

\textbf{Step 3:} Determine the set of patch-wise mutual top-$k$ correspondences $\mathcal{M}'$ using Eq.~(\ref{eq11}) and Eq.~(\ref{eq12})  \;

\textbf{Step 4:} Calculate the patch-level similarity vector as described in Eq.~(\ref{eq13}) and Eq.~(\ref{eq_weight})\;

\textbf{Step 5:} Derive the final similarity score $\Psi$ by Eq.~(\ref{eq14}) \;
\Return{$\Psi$}
\label{alg1}
\end{algorithm}

\noindent\textbf{Top-$k$ Filtering.}  
Since the patches are obtained by fixed partitioning, the semantic content within one patch may correspond to multiple patches in the other modality. To address this, we introduce a Top-$k$ filtering strategy. Formally, given the refined feature embeddings $\mathbf{F}_{v}^{\prime}, \mathbf{F}_{i}^{\prime} \in \mathbb{R}^{N \times C}$ for visible and infrared images, we first compute the bi-directional patch-wise similarity matrices:
\begin{equation}
    \mathbf{S}_{vi} = \texttt{s}\left( \frac{\mathbf{F}_{v}^{\prime} \mathbf{F}_{i}^{\prime T}}{\sqrt{C}} \right),
    \mathbf{S}_{iv} = \texttt{s}\left( \frac{\mathbf{F}_{i}^{\prime} \mathbf{F}_{v}^{\prime T}}{\sqrt{C}} \right),
\label{eq9}
\end{equation}
where $\texttt{s}(\cdot)$ denotes the softmax function, which is applied along the row dimension for each matrices. Although $\mathbf{S}_{iv}$ is the transpose of $\mathbf{S}_{vi}$ in raw dot-product form, the row-wise normalization renders them directionally distinct. 

As shown in Fig.~\ref{fig2}, given patch-wise similarity matrices $\mathbf{S}_{vi}, \mathbf{S}_{iv} $ between visible and infrared images, we first perform top-$k$ filtering to retain most responsive candidates and construct the neighborhood sub-graph for each patch:
\begin{equation}
    \mathbf{R}_{v - i} = \texttt{TopK}(\mathbf{S}_{vi}, k), \mathbf{R}_{i - v} =\texttt{TopK}(\mathbf{S}_{iv}, k),
\label{eq10}
\end{equation}
where $\mathbf{R}_{v - i}$ stores the top-$k$ infrared patches for each visible patch, and vice versa for $\mathbf{R}_{i - v}$. Due to the directional nature of the similarity matrices, the neighborhood relations are generally non-reciprocal, \textit{i.e.}, a patch included in another patch’s neighborhood may not include the latter in its own. (\textit{e.g.}, the $\mathbf{R}_{v - i}[0]$ and $\mathbf{R}_{i - v}[0]$ in the middle of Fig.~\ref{fig2}).

\noindent\textbf{Reciprocal Patch Mining (RPM).}  
In the infrared modality, the absence of color information reduces the distinguishability of different body regions, making one-sided patch matching prone to introducing noise. To capture reliable local correspondences across modalities while suppressing one-sided or noisy matches, we introduce a Reciprocal Patch Mining strategy that progressively filters and completes patch-level interactions. We first extract the mutually selected patch pairs to construct the mutual match set:
\begin{equation}
\mathcal{M} = \{(p, q) \mid q \in \mathbf{R}_{v - i}[p] \ \text{and} \ p \in \mathbf{R}_{i - v}[q] \},
\label{eq11}
\end{equation}
after RPM, some visible patches may not be selected in any reciprocal pair and would be excluded from the final scoring. To ensure all visible patches are involved, we extend mutual match set to a robust version via smooth complementary: 
\begin{equation}
\mathcal{M}' = \mathcal{M} \cup \{(p, q^*) \mid \mathcal{M}_p = \emptyset,\ q^* = \arg\max \mathbf{S}_{vi}[p] \},
\label{eq12}
\end{equation}
where $\mathcal{M}_p = \{q \mid (p, q) \in \mathcal{M} \}$ denotes the mutual matches of patch $p$. This strategy guarantees full coverage while preserving the reliability of reciprocal selection.

We then compute a patch-level similarity score vector $\hat{\mathbf{S}} \in \mathbb{R}^{N}$ for each visible image by averaging the similarities over its matched infrared patches:
\begin{equation}
\hat{\mathbf{S}}[p] = \frac{1}{|\mathcal{M}'_p|} \sum_{q \in \mathcal{M}'_p} w_{p,q}\cdot\mathbf{S}_{vi}[p, q],\mathcal{M}'_p = \{q \mid (p, q) \in \mathcal{M}' \},
\label{eq13}
\end{equation}
\begin{equation}
w_{p,q} = \begin{cases}
1, & \text{if } (p, q) \in \mathcal{M} \\
\alpha, & \text{otherwise}
\end{cases}
\label{eq_weight}
\end{equation}
Here, $\alpha$ is the penalty coefficient that suppresses the contribution of non-mutual matches introduced via smooth complementary selection.

% The resulting vector $\hat{\mathbf{S}} \in \mathbb{R}^{N}$ captures reliability-aware patch-level matching cues, which are later aggregated for final scoring.

\noindent\textbf{Adaptive Patch Aggregation via Confidence Modeling.}  
To convert the similarity vector $\hat{\mathbf{S}}$ into a final image-level similarity score, we design a lightweight yet expressive prediction head, named \textbf{Confidence-Aware Scoring Module (CASM)}. This module, implemented as a small MLP, adaptively learns to weight different patch-level cues based on their informativeness and context:
\begin{equation}
    \Psi = \sigma(\texttt{CASM}(\hat{\mathbf{S}})),
\label{eq14}
\end{equation}
where $\Psi$ denotes the final similarity score between the input pair, and $\sigma(\cdot)$ represents the Sigmoid activation function.

This design is motivated by the observation that not all mutual matches contribute equally, where some correspond to salient body parts or distinctive accessories, while others may introduce noise. By learning a soft aggregation scheme over patch-level responses, the CASM enables the model to prioritize informative matches and suppress misleading ones, enhancing both robustness and discriminability.

\begin{table*}[t]
\small
\centering
\caption{Comparison with state-of-the-art methods on SYSU-MM01. $^R$ means using Re-Ranking. The best and second-best results are highlighted in bold and underlined, respectively.}
\renewcommand{\arraystretch}{1}
\renewcommand\tabcolsep{10pt}
	\begin{tabular}{c|c|cc|cc|cc|cc}
		\hline
		\multicolumn{1}{c|}{\multirow{3}{*}{Model}} & \multicolumn{1}{c|}{\multirow{3}{*}{Venue}}& \multicolumn{4}{c|}{All-Search} &\multicolumn{4}{c}{Indoor-Search}\\ \cline{3-10}
              &&\multicolumn{2}{c|}{ Single-Shot} &\multicolumn{2}{c|}{ Multi-Shot} &\multicolumn{2}{c|}{ Single-Shot} &\multicolumn{2}{c}{ Multi-Shot}\\ \cline{3-10}
              && Rank-1 & mAP & Rank-1 & mAP & Rank-1 & mAP & Rank-1 & mAP \\
\hline
NFS \cite{c26} &CVPR'21 & 56.91 & 55.45 & 63.51 & 48.56 & 62.79 & 69.79 & 70.03 & 61.45 \\
cm-SSFT \cite{c27}&CVPR'20 & 61.60 & 63.20 & 63.40 & 62.00 & 70.50 & 72.60 & 73.00 & 72.40 \\
MID \cite{c35}&AAAI'22 & 60.27 & 59.40 & - & - & 64.86 & 70.12 & - & - \\
CMTR \cite{c28}&TMM'23 & 62.58 & 61.33 & 68.39 & 55.69 & 67.02 & 73.78 & 75.40 & 66.84 \\
MCLNet \cite{c29}&ICCV'21 & 65.40 & 61.98 & - & - & 72.56 & 76.58 & - & - \\
MPANet \cite{c37}&CVPR'21 & 70.58 & 68.24 & 75.58 & 62.91 & 76.74 & 80.95 & 84.22 & 75.11 \\
MAUM \cite{c30}&CVPR'22 & 71.68 & 68.79 & - & - & 76.97 & 81.94 & - & - \\
FMCNet \cite{c36}&CVPR'22 & 66.34 & 62.51 & 73.44 & 56.06 & 68.15 & 74.09 & 78.86 & 63.82 \\
CMT \cite{c38}&ECCV'22 & 71.88 & 68.57 & 80.23 & 63.13 & 76.90 & 79.91 & 84.87 & 74.11 \\
CAL \cite{c31}&ICCV'23 & 74.66 & 71.73 & 77.05 & 64.86 & 79.69 & 83.68 & 86.97 & 78.51 \\
SAAI \cite{c32}&ICCV'23& 75.90 & \underline{77.03} & 82.86 & \underline{82.39} & 83.20 & 88.01 & 90.73 & \underline{91.30} \\
SEFL \cite{c33}&CVPR'23& 77.12 & 72.33 & - & - & 82.07 & 82.95 & - & - \\
PartMix \cite{c34}&CVPR'23& 77.78 & 74.62 & 80.53 & 69.84 & 81.52 & 84.38 & 87.99 & 79.95 \\
PMT \cite{c47}&AAAI'23& 67.53 & 64.98 & - & - & 71.66 & 76.52 & - & - \\
protoHPE \cite{c39}&ACMMM'23& 71.92 & 70.59 & - & - & 77.81 & 81.31 & - & - \\
MUN \cite{c40}&ICCV'23& 76.24 & 73.81 & - & - & 79.42 & 82.06 & - & - \\
MSCLNet \cite{c41}&ECCV'22& 76.99 & 71.64 & - & - & 78.49 & 81.17 & - & - \\
DEEN \cite{c15}&CVPR'23& 74.70 & 71.80 & - & - & 80.30 & 83.30 & - & - \\
HOS-Net \cite{c24}&AAAI'24& 75.60 & 74.20 & - & - & 84.20 & 86.70 & - & - \\
WRIM-Net \cite{c25}&ECCV'24& 77.40 & 75.40 & \underline{83.20}  & 71.10 & \underline{86.20}  & \underline{88.10} & \underline{92.10}  & 84.60 \\
DiVE \cite{c74}&AAAI'25& \underline{79.07} & 74.96 & -  & - & 82.98  & 85.90 & -  & - \\
\rowcolor{gray!20}
BIT (ours)&CVPR'26& \textbf{80.53} & \textbf{79.76} & \textbf{85.31} & \textbf{84.37} & \textbf{87.42} & \textbf{89.25} & \textbf{93.10} & \textbf{91.41} \\
\hline
CIFT$^R$ \cite{c43}&ECCV'22& 74.08 & 74.79 & 79.74 & 75.56 & 81.82 & 85.61 & 88.32 & 86.42 \\
IDKL$^R$ \cite{c44}&CVPR'24& 81.42 & 79.85 & 84.34 & 78.22 & 87.14 & 89.37 & 94.30 & 88.75 \\
\rowcolor{gray!20}
BIT (ours)$^R$&CVPR'26& \textbf{84.42} & \textbf{83.64} & \textbf{87.31} & \textbf{84.42} & \textbf{91.64} & \textbf{90.32} & \textbf{95.74} & \textbf{91.31} \\
\hline
	\end{tabular}

\label{tab:sys}
\end{table*}

\subsection{Training}

To effectively optimize both the backbone and our proposed BIT, we adopt a two-stage training paradigm.

In the first stage, we train the backbone independently using the standard modality-invariant loss $\mathcal{L}_{base}$, which consists of the commonly used identity classification loss and triplet loss. This stage aims to learn discriminative and robust feature embeddings for both visible and infrared modalities, while preventing premature overfitting from modality-specific interactions. The BIT is not involved in this phase.
 
In the second stage, the backbone parameters are frozen, and we exclusively optimize the BIT to enhance cross-modality matching. Given the final similarity score $\Psi \in [0, 1]$ predicted by the QA scoring module, and the ground-truth label $y \in \{0, 1\}$ indicating whether the input pair belongs to the same identity, the pairwise matching loss is:
\begin{equation}
    \mathcal{L}_{pair} = - \left( y \cdot \log(\Psi) + (1 - y) \cdot \log(1 - \Psi) \right),
    \label{eq:pair}
\end{equation}
% This binary cross-entropy loss encourages the model to produce higher scores for positive pairs and suppress similarities between negative pairs, effectively reinforcing identity-discriminative patch correspondences. 
The total training objective in the second stage is the weighted sum of the two loss terms:
\begin{equation}
    \mathcal{L}_{\text{total}} =  \mathcal{L}_{pair} + \lambda \cdot\mathcal{L}_{AC},
\label{eq15}
\end{equation}
where $\lambda$ is a balancing hyperparameter. This two-stage training ensures that the model captures strong modality-invariant representations before refining its cross-modality matching capacity.

\section{Experiments} 
\subsection{Datasets and Evaluation Metrics}

% \textbf{SYSU-MM01} is the largest and most widely used benchmark for VI-ReID. It contains 287,628 RGB images and 15,792 IR images from 491 identities. The dataset is split into a training set with 395 identities (22,258 RGB and 11,909 IR images) and a testing set with the remaining 96 identities. The testing set includes 3,803 infrared query images and 301 visible gallery images. Evaluation is conducted under two standard protocols: \textit{all-search mode}, which considers both indoor and outdoor gallery images, and \textit{indoor-search mode}, which includes only indoor gallery images. All evaluations are performed under both \textit{single-shot} and \textit{multi-shot} setting, and results are averaged over 10 random trials.

% \textbf{RegDB} comprises 8,240 images of 412 identities, captured by paired visible and infrared cameras. Each identity has 10 visible and 10 infrared images. The dataset is divided equally for training and testing. Two cross-modality matching settings are evaluated: visible-to-infrared (V2I) and infrared-to-visible (I2V), where queries and galleries belong to different modalities.

% \textbf{LLCM} is a large scale dataset designed for VI-ReID. It is split into training and testing sets at a 2:1 ratio. Similar to RegDB, LLCM supports both V2I and I2V matching scenarios, providing a more realistic and challenging environment for evaluating cross-modality robustness under adverse illumination.

\textbf{Datasets.} We evaluate the proposed BIT on three widely used VI-ReID benchmarks: SYSU-MM01, LLCM, and RegDB. Detailed dataset statistics and evaluation protocols are provided in \textbf{Supplementary Material}. In addition, we conduct controlled data-imbalance experiments by manually reducing the number of training samples per identity to assess BIT’s robustness under varying degrees of data imbalance. The corresponding settings and experimental results are reported in \textbf{Supplementary Material}.

\noindent\textbf{Evaluation Metrics.}
We leverage two widely used metrics for person ReID: the Cumulative Matching Characteristic (CMC) and mean Average Precision (mAP).

% CMC evaluates the rank-based retrieval accuracy, while mAP measures overall retrieval performance by considering both precision and recall across all ranks.

\begin{table}[t]
\small
    \centering
    \caption{Comparison with SOTA models on the LLCM dataset.}
    \renewcommand{\arraystretch}{1.1}
    \renewcommand\tabcolsep{9pt}
    \begin{tabular}{c|c|c|c|c}
        \hline
         \multicolumn{1}{c|}{\multirow{2}{*}{Model}}& \multicolumn{2}{c|}{Visible2Infrared} &\multicolumn{2}{c}{Infrared2Visible} \\ \cline{2-5}
          & Rank-1 & mAP & Rank-1 & mAP \\ \cline{1-5}
          DDAG \cite{c46}  & 40.3 & 48.4 & 48.0 & 52.3 \\
          CAJ \cite{c45} & 56.5 & 59.8 & 48.8 & 56.6 \\
          DEEN \cite{c15}  & 62.5 & 65.8 & 54.9 & 62.9 \\
          HOS-Net \cite{c24}  & \underline{64.9} & \underline{67.9} & \underline{56.4} & \underline{63.2} \\
           \cline{1-5}\rowcolor{gray!20}
          BIT (ours) & \textbf{73.1} & \textbf{69.2} & \textbf{66.7} & \textbf{67.2} \\ 
          \cline{1-5}
    \end{tabular}       
    \label{tab:llcm}    
    
\end{table}
\begin{table}
\small
    \centering
    \caption{Comparison with SOTA models on RegDB dataset.}
    \renewcommand{\arraystretch}{1.1}
    \renewcommand\tabcolsep{10pt}
    \begin{tabular}{c|c|c|c|c}
        \hline
         \multicolumn{1}{c|}{\multirow{2}{*}{Model}}& \multicolumn{2}{c|}{Visible2Infrared} &\multicolumn{2}{c}{Infrared2Visible} \\ \cline{2-5}
          & Rank-1 & mAP & Rank-1 & mAP \\ \cline{1-5}
          PMT & 84.83 & 76.55 & 84.16 & 75.13 \\
          MAUM & 87.87 & 85.09 & 86.95 & 84.34 \\
          FMCNet & 89.12 & 84.43 & 88.38 & 83.86 \\
          SAAI & 91.07 & 91.45 & 92.09 & \underline{92.01} \\
          CMT & 95.17 & 87.30 & 91.97 & 84.46 \\
          MUN & \underline{95.19} & 87.15 & 91.86 & 85.01 \\
          DEEN & 91.10 & 85.10 & 89.50 & 83.40 \\
          HOS-Net & 94.70 & 90.04 & 93.30 & 89.20 \\
          \cline{1-5}
          CIFT$^R$ & 91.96 & \underline{92.00} & 90.30 & 90.78 \\
          IDKL$^R$ & 94.72 & 90.19 & \textbf{94.22}& 90.43 \\
          \cline{1-5}\rowcolor{gray!20}
          BIT (ours) & \textbf{96.12} & \textbf{92.43} & \underline{93.65} & \textbf{93.54} \\
          \cline{1-5}
    \end{tabular}     
    \label{tab:regdb} 
\end{table}
\begin{table}[t]
\caption{Ablation studies of each component on SYSU-MM01 (Single-Shot All-Search mode).}
\centering
\small
\renewcommand{\arraystretch}{1.15}
\setlength{\tabcolsep}{5pt}
\begin{tabular}{c|c|c|c|c|cc}
\hline
\multirow{2}{*}{Index} & \multirow{2}{*}{Base} & \multirow{2}{*}{BCI} & \multirow{2}{*}{$\mathcal{L}_{AC}$} & \multirow{2}{*}{QA Scoring} & \multicolumn{2}{c}{SYSU-MM01} \\ \cline{6-7}
                       &                       &                      &                                    &                             & Rank-1 & mAP \\ \hline
1 & \ding{52} & \ding{55} & \ding{55} & \ding{55} & 69.23 & 66.02 \\
2 & \ding{52} & \ding{52} & \ding{55} & \ding{55} & 75.24 & 73.35 \\
3 & \ding{52} & \ding{52} & \ding{52} & \ding{55} & 76.42 & 74.54 \\
4 & \ding{52} & \ding{52} & \ding{55} & \ding{52} & 79.53 & 79.02 \\
5 & \ding{52} & \ding{52} & \ding{52} & \ding{52} & \textbf{80.53} & \textbf{79.76} \\ \hline
\end{tabular}
\label{ablation}
\end{table}

\subsection{Implementation Overview}
All experiments are conducted on a single NVIDIA L20 GPU using PyTorch. We follow the same backbone configuration as PMT~\cite{c47}, employing a ViT-B/16 \cite{c48} model pre-trained on ImageNet \cite{c42}. The number of stacked BCI blocks $T$ is set to 3. To balance computational efficiency and representation quality, the overlap stride is set to 12.

\noindent\textbf{Hyperparameters.}  
The Top-$k$ parameter in Eq.~(\ref{eq10}) is set to 3, the penalty coefficient $\alpha$ in Eq.~(\ref{eq_weight}) is set to 0.20, and the balancing weight $\lambda$ in Eq.~(\ref{eq15}) is set to 0.6,  both selected via grid search.

\subsection{Comparison with State-of-the-Art Methods}
We compare the proposed BIT with several state-of-the-art methods on the SYSU-MM01, LLCM and RegDB. The results are presented in Tab.~\ref{tab:sys}, Tab.~\ref{tab:llcm}, and Tab.~\ref{tab:regdb}, with the best results highlighted in bold and the second-best results are highlighted underlined. For fair comparison, we apply re-ranking~\cite{c50} only when comparing with prior works that also it and report results without re-ranking for the others.

\noindent\textbf{SYSU-MM01.} Tab.~\ref{tab:sys} compares BIT with existing methods on the SYSU-MM01 dataset. Our method consistently outperforms all SOTA approaches under both Single-Shot and Multi-Shot settings in All-Search and Indoor-Search scenarios. Without using re-ranking, BIT achieves remarkable performance in the All-Search Single-Shot setting, reaching the highest \textbf{80.53\%} Rank-1 and \textbf{79.76\%} mAP among all methods. In the Indoor-Search setting, BIT also achieves the best results, with \textbf{87.42\%} Rank-1 and \textbf{89.25\%} mAP. 

% When incorporating re-ranking, BIT further boosts its performance, achieving \textbf{84.42\%} Rank-1 and \textbf{92.61\%} mAP in the All-Search Single-Shot setting, surpassing all methods by a clear margin.

\noindent\textbf{LLCM.} As shown in Tab.~\ref{tab:llcm}, BIT surpasses all competitors on the LLCM dataset for both V2I and I2V tasks. It achieves the best Rank-1 score of \textbf{73.1\%} and the best mAP of \textbf{69.2\%} in V2I retrieval. For I2V retrieval, BIT obtains both the highest Rank-1 (\textbf{66.7\%}) and mAP (\textbf{67.2\%}) scores. 

\noindent\textbf{RegDB.} Tab.~\ref{tab:regdb} presents the performance comparison on the RegDB dataset. BIT achieves new SOTA results in both V2I and I2V retrieval tasks. Without re-ranking, BIT attains an outstanding \textbf{96.12\%} Rank-1 and \textbf{92.43\%} mAP in the V2I task, significantly outperforming prior methods. In the I2V task, BIT also achieves a competitive Rank-1 of \textbf{93.65\%} and the best mAP of \textbf{93.54\%}.

% In summary, our BIT consistently achieves top performance, leading all eight metrics on SYSU-MM01, three of four on RegDB, and all four on LLCM without relying on re-ranking, clearly demonstrating its superiority.

\begin{table}
\caption{The influence of Bi-directional design on the performance of SYSU-MM01(Single-Shot mode).}
\centering
\small
\renewcommand{\arraystretch}{1.15}
\setlength{\tabcolsep}{5pt}
\begin{tabular}{c|cc|cc}
\hline
\multirow{2}{*}{Model} &  \multicolumn{2}{c|}{All-Search} & \multicolumn{2}{c}{Indoor-Search} \\ \cline{2-5}
             &         Rank-1                  &           mAP                  & Rank-1 & mAP \\ \hline
Baseline & 69.23 & 66.02 & 72.10 & 77.32 \\
+ Cross-Attn & 68.68 & 64.67 & 70.23 & 74.34 \\
\rowcolor{gray!20} + BCI & \textcolor{red}{75.24} & \textcolor{red}{73.35} & \textcolor{red}{78.43} & \textcolor{red}{82.36} \\
\hline
\end{tabular}
\label{tab5}
\end{table}
\subsection{Ablation Studies}
We conduct ablation studies on the All-Search Single-Shot mode in SYSU-MM01 to evaluate the effectiveness of our core components: the BCI for early-stage modality interaction, the Aggregation Contrastive Loss $\mathcal{L}_{AC}$ for cross-modalities interaction constraint and the QA Scoring module for dynamic patch matching. We use a reproduced PMT as our baseline, and the results are reported in Tab.~\ref{ablation}.

\noindent\textbf{Effectiveness of BCI.} Introducing BCI leads to a significant performance gain (+6.01\% Rank-1, +7.33\% mAP), demonstrating that early cross-modal feature interaction is crucial in reducing modality discrepancy in VI-ReID.

\noindent \textbf{The influence of the Bi-directional design.} Comparing with standard cross-attention, BCI adopts a novel bi-directional design that enables mutual feature compensation between modalities, which can produce better aligned intermediate features for subsequent matching. To validate it, we implement a variant that the baseline PMT incorporating standard cross-attention with pairwise loss $\mathcal{L}_{pair}$ (as in ALBEF \cite{c80}). As shown in Tab.~\ref{tab5}, the variant with standard cross-attention even performs worse than the baseline, while our BCI brings significant gains, highlighting the superiority of the bi-directional design.

\noindent\textbf{Effectiveness of $\mathcal{L}_{AC}$.} Adding the aggregation contrastive loss further improves both Rank-1 and mAP, showing its effectiveness in enhancing feature alignment by adaptively mining informative positive pairs across modalities.

\noindent\textbf{Effectiveness of QA Scoring.} The QA Scoring module yields notable improvements (+4.11\% Rank-1, +5.22\% mAP). This highlights its ability to achieve more robust and reliable local correspondences.

% \noindent\textbf{Full Model.} Combining all components results in the highest performance (\textbf{80.53\%} Rank-1, \textbf{79.76\%} mAP), confirming their effectiveness in VI-ReID.

\begin{figure}[t]
  \centering
  
   \includegraphics[width=1.0\linewidth]{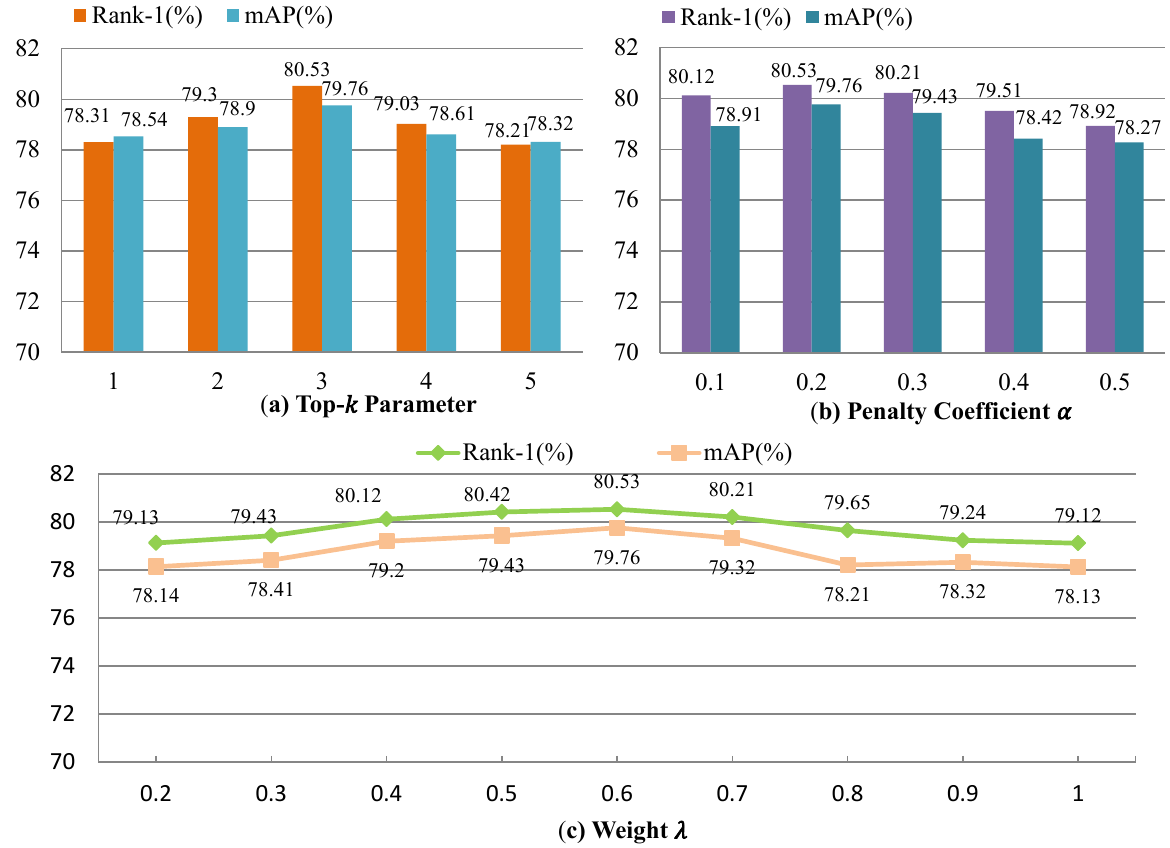}

   \caption{The impact of Top-$k$ parameter in Eq.~(\ref{eq10}), penalty coefficient $\alpha$ in Eq.~(\ref{eq_weight}) and balancing weight $\lambda$ in Eq.~(\ref{eq15}).}
   \label{Fig3}
\end{figure}

\subsection{Model Analysis}
\textbf{Parameter Analysis.} We conduct detailed parameter analysis experiments under the All-Search Single-Shot setting on the SYSU-MM01 dataset. We investigate the influence of the Top-$k$ parameter in Eq.~(\ref{eq10}), penalty coefficient $\alpha$ in Eq.~(\ref{eq_weight}) and the balancing weight $\lambda$ in Eq.~(\ref{eq15}). As shown in Fig.~(\ref{Fig3}), both Rank-1 and mAP performance improve as $k$ increases and peak when $k=3$. In addition, we find that the best performance is achieved when the penalty coefficient $\alpha$ is set to 0.2. As for $\lambda$, the optimal performance is obtained when $\lambda$ is set to 0.6.

\begin{figure}[t]
  \centering
   \includegraphics[width=1.0\linewidth]{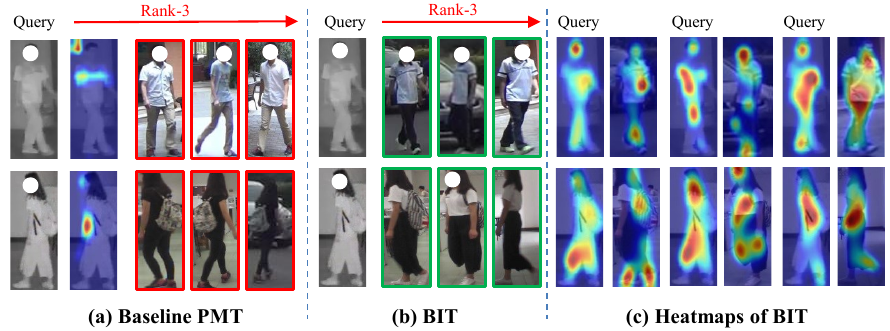}
   \caption{Qualitative comparison of retrieval results and attention maps. (a) Rank-3 retrieval results of baseline along with query’s attention maps. (b) Rank-3 retrieval results of BIT. (c) Attention maps of BIT for different gallery images.}
   \label{Fig5}
\end{figure}

\noindent\textbf{Visualization Analysis.} To qualitatively demonstrate the effectiveness of BIT, we present retrieval results of BIT compared with the baseline in Fig.~\ref{Fig5}. Overall, BIT can effectively improve the ranking results with more correct matched images than the baseline. 
We further visualize the attention maps of the baseline in the second row of Fig.\ref{Fig5}(a) and those of BIT in Fig.\ref{Fig5}(c). While the baseline exhibits fixed and noisy attention, BIT focuses more accurately on pedestrian regions and suppresses background noise. Moreover, BIT adaptively adjusts its attention based on the gallery. For instance, in the second column of the sixth row, when the image depicts a girl without a bag, BIT correctly reduces attention to the bag in the query, demonstrating its ability to perform adaptive matching.

\begin{figure}[t]
\centering
\includegraphics[width=0.95\columnwidth]{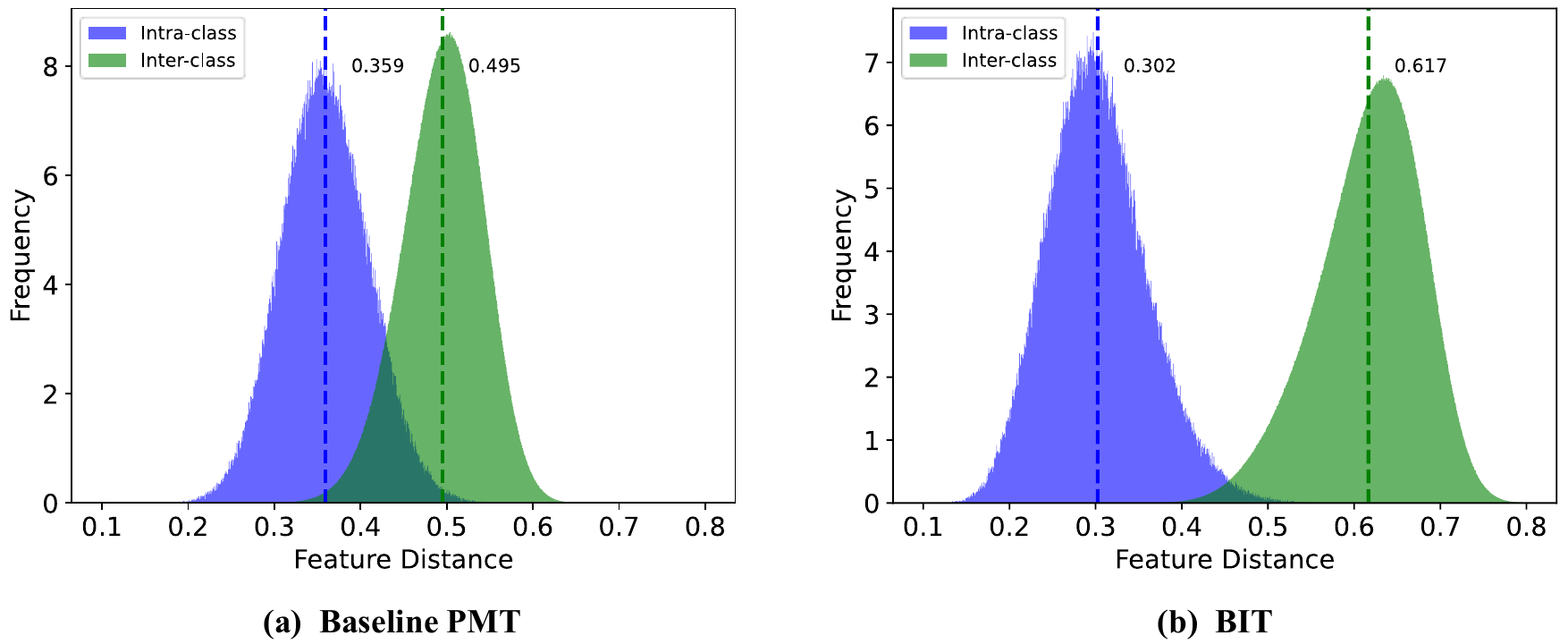}
\caption{Similarity distributions of positive and negative pairs on SYSU-MM01 during inference.}
\label{fig:supp1}
\end{figure}
\noindent\textbf{Similarity Distribution.} Fig.~\ref{fig:supp1} illustrates the similarity distributions of positive and negative pairs on the SYSU-MM01 dataset during inference. To ensure a fair comparison, we normalize all distances to the [0, 1] range. Specifically, BIT directly computes similarity (bounded between 0 and 1), while the baseline PMT \cite{c47} first derives cosine distances in the [0, 2] range, which are subsequently normalized. Comparing with the baseline, BIT decreases the mean similarity of positive pairs from 0.359 to 0.302, while increasing that of negative pairs from 0.495 to 0.617.

\noindent\textbf{Complexity Analysis.} While BIT incurs a quadratic $\mathcal{O}(N^2)$ cost during training, its inference is substantially more efficient with a coarse candidate inference strategy. Specifically, preliminary features extracted by the backbone are first used to compute coarse similarity scores for all query–gallery pairs, from which the Top-$K$ candidates are selected for each query. The final matching scores of these candidates are then computed using the BCI module and QA Scoring module. Thus, the inference complexity is $\mathcal{O}(1)$ for feature extraction and $\mathcal{O}(K)$ for matching, making the inference overhead independent of the gallery size. Notably, $K$ can be fixed to a small constant ($K=50$) while yielding identical performance. In this setting, BIT achieves \textbf{14ms} latency per query, compared to \textbf{10ms} for PMT.

% Since gallery features can be cached offline, each query requires only $\mathcal{O}(1)$ for feature extraction and $\mathcal{O}(M)$ for patch-wise matching, ensuring practical scalability. Importantly, Pairwise matching is implemented efficiently via batched GPU operations, enabling scalable retrieval with negligible computational overhead.
% To evaluate practical efficiency, we conduct inference time analysis on SYSU-MM01 with a unified batch size of 256. As PMT serves as the baseline of our BIT, we adopt it for comparison to ensure consistency. In this setting, BIT processes query-gallery pairs while PMT handles 256 queries. BIT achieves \textit{\textbf{21ms}} per query compared to \textit{\textbf{10ms}} for PMT. Despite the slight increase in latency, BIT remains efficient enough for deployment and delivers significantly higher accuracy.

\section{Conclusion}
We propose a novel matching-based Bi-directional Interaction Transformation (BIT) network for VI-ReID, which leverages a Bi-directional Cross interaction (BCI) module to enable adaptive feature compensation between modalities and a Query Aware Scoring (QA Scoring) module to aggregates reliable patch correspondences while suppressing noisy matches. By introducing this non-rigid interaction mechanism, BIT effectively bridges the modality gap, even under severe data imbalance. Extensive experiments demonstrate that BIT consistently outperforming existing rigid feature learning approaches by a significant margin.
\clearpage
{
    \small
    \bibliographystyle{ieeenat_fullname}
    \bibliography{main}
}
\clearpage
\setcounter{page}{1}
\maketitlesupplementary

\section{Dataset}
We evaluate the BIT on three widely-used VI-ReID benchmarks: SYSU-MM01, LLCM and RegDB.

\textbf{SYSU-MM01} \cite{c14} is the largest and most widely used benchmark for VI-ReID. It contains 287,628 RGB images and 15,792 IR images from 491 identities. The dataset is split into a training set with 395 identities (22,258 RGB and 11,909 IR images) and a testing set with the remaining 96 identities. The testing set includes 3,803 infrared query images and 301 visible gallery images. Evaluation is conducted under two standard protocols: \textit{all-search mode}, which considers both indoor and outdoor gallery images, and \textit{indoor-search mode}, which includes only indoor gallery images. All evaluations are performed under both \textit{single-shot} and \textit{multi-shot} setting, and results are averaged over 10 random trials.

\textbf{RegDB} \cite{c49} comprises 8,240 images of 412 identities, captured by paired visible and infrared cameras. Each identity has 10 visible and 10 infrared images. The dataset is divided equally for training and testing. Two cross-modality matching settings are evaluated: visible-to-infrared (V2I) and infrared-to-visible (I2V), where queries and galleries belong to different modalities.

\textbf{LLCM} \cite{c15} is a large scale dataset designed for VI-ReID. It is split into training and testing sets at a 2:1 ratio. Similar to RegDB, LLCM supports both V2I and I2V matching scenarios, providing a more realistic and challenging environment for evaluating cross-modality robustness under adverse illumination.

\begin{table}[h]
\centering
  \renewcommand{\arraystretch}{0.8}
\fontsize{9pt}{13pt}\selectfont
\caption{Dataset statistics of RegDB, SYSU-MM01 and LLCM.}\label{tab1:tab1}
\begin{tabular}{l|ccccc}
\hline
Datasets                     & IDs   & Images & VIS / IR cam  \\
\hline
RegDB & 412   & 8,240  & 1 / 1            \\   
SYSU-MM01   & 491   & 287,628 & 4 / 2         \\   
LLCM                 & 1,064 & 46,767 & 9 / 9  \\ \hline
\end{tabular}%}
  \vspace{-0.2cm}
  \vspace{-0.5cm}
  \label{tab:tab1}
\end{table}

\section{Implementation Details}

All experiments are conducted on a single NVIDIA L20 GPU using PyTorch. We follow the same backbone configuration as PMT \cite{c47}, employing a ViT-B/16 \cite{c48} model pre-trained on ImageNet \cite{c42}. The number of stacked BCI blocks $T$ is set to 3. To balance computational efficiency and representation quality, the overlap stride is set to 12. All person images are resized to $256 \times 128$ before being fed into the network. For data augmentation, we apply random horizontal flipping and random erasing to both modalities. Additionally, color jitter and Gaussian blur are applied exclusively to infrared images to enhance modality robustness.

\textbf{Stage 1: Backbone Training.}  
In the first training stage, we train the backbone to obtain strong modality-invariant representations. Each mini-batch contains 64 images sampled from 8 identities, with 4 visible and 4 infrared images per identity. The model is optimized using AdamW with a cosine annealing learning rate scheduler. The initial learning rate is set to $3 \times 10^{-4}$, and the weight decay is set to $1 \times 10^{-4}$. This stage is trained for 40 epochs on all datasets.

\textbf{Stage 2: BIT Training.}  
In the second stage, the backbone network is frozen and only the proposed BIT modules are optimized. Each mini-batch is constructed from 4 randomly sampled identities, each contributing 4 visible and 4 infrared images, resulting in 32 images per batch. Unlike traditional methods that operate on individual images, BIT takes visible-infrared feature pairs as input. This design yields 256 cross-modal pairs per batch, effectively transforming a batch of $B$ images into $B^2/4$ training samples. We continue to use AdamW with the same learning rate and scheduler settings. This stage is trained for 60 epochs.

\section{Controlled Data-Imbalance Experiments}
\begin{table*}[t]
\small
\centering
\caption{Comparison with SOTA methods on the SYSU-MM01 dataset under controlled Data-Imbalance settings. The best are highlighted in bold. $^\dagger$ denotes our reproduced results.}
\renewcommand{\arraystretch}{1}
\renewcommand\tabcolsep{10pt}
\begin{tabular}{c|c|cc|cc}
\hline
\multicolumn{1}{c|}{\multirow{2}{*}{Settings}} & 
\multicolumn{1}{c|}{\multirow{2}{*}{Model}} &
\multicolumn{2}{c|}{All-Search } &
\multicolumn{2}{c}{Indoor-Search} \\ \cline{3-6}
& & Rank-1 & mAP & Rank-1 & mAP \\
\hline

\multirow{7}{*}{Original}
&MCLNet \cite{c29} & 65.40 & 61.98 & 72.56 & 76.58 \\
&MPANet \cite{c37} & 70.58 & 68.24 & 76.74 & 80.95 \\
&SAAI \cite{c32} & 75.90 & 77.03 & 83.20 & 88.01 \\
&DEEN \cite{c15} & 74.70 & 71.80 & 80.30 & 83.30 \\
&HOS-Net \cite{c24} & 75.60 & 74.20 & 84.20 & 86.70 \\
\rowcolor{gray!20}
&PMT (baseline)$^\dagger$ \cite{c47} & 69.23 & 66.02 & 73.24 & 77.91 \\
\rowcolor{gray!20}
&BIT (ours) & \textbf{80.53} (+11.3) & \textbf{79.76} (+13.74) & \textbf{87.42}(+14.18) & \textbf{89.25} (+11.28) \\
\hline

\multirow{7}{*}{10\% Inf Reduce}
&MCLNet & 60.31 & 57.46 & 67.31 & 73.42 \\
&MPANet & 66.34 & 64.21 & 72.67 & 76.47 \\
&SAAI & 71.43 & 72.97 & 78.43 & 83.42 \\
&DEEN & 70.21 & 65.32 & 76.74 & 78.95 \\
&HOS-Net & 70.45 & 69.98 & 80.26 & 81.78 \\
\rowcolor{gray!20}&PMT(baseline) & 65.20 & 62.43 & 70.45 & 72.54 \\
\rowcolor{gray!20}
&BIT (ours) & \textbf{79.67}(+14.47) & \textbf{79.39}(+16.96) & \textbf{86.17}(+15.72) & \textbf{87.96}(+15.42) \\
\hline

\multirow{7}{*}{20\% Inf Reduce}
&MCLNet & 56.42 & 53.86 & 63.52 & 69.37 \\
&MPANet & 62.67 & 60.93 & 69.42 & 72.81 \\
&SAAI & 66.57 & 68.91 & 73.78& 77.92 \\
&DEEN & 65.76 & 61.34 & 72.89 & 72.79 \\
&HOS-Net & 64.70& 64.71& 75.54 & 76.47 \\
\rowcolor{gray!20}&PMT(baseline) & 60.40 & 58.37 & 66.87 & 69.65 \\
\rowcolor{gray!20}
&BIT (ours) & \textbf{77.80}(+17.40) & \textbf{78.65}(+20.28) & \textbf{85.45}(+18.58) & \textbf{86.47}(+16.82) \\
\hline

\multirow{7}{*}{10\% Vis Reduce}
&MCLNet & 62.79 & 59.86 & 70.47 & 74.89 \\
&MPANet & 68.92 & 67.31 & 73.79 & 78.06 \\
&SAAI & 74.32 & 74.97 & 80.43 & 85.76 \\
&DEEN & 73.07 & 68.92 & 77.87 & 80.25 \\
&HOS-Net & 72.76 & 72.10 & 82.36 & 82.95 \\
\rowcolor{gray!20}&PMT(baseline) & 67.16 & 63.92 & 71.64 & 73.17 \\
\rowcolor{gray!20}
&BIT (ours) & \textbf{79.51}(+12.35) & \textbf{79.42}(+15.5) & \textbf{86.65}(+15.01) & \textbf{88.21}(+15.04) \\
\hline

\multirow{7}{*}{20\% Vis Reduce}
&MCLNet & 59.82 & 57.93 & 66.82 & 70.60 \\
&MPANet & 65.78 & 64.42 & 69.75 & 75.32 \\
&SAAI & 70.24 & 71.66& 77.81 & 81.62 \\
&DEEN & 69.65 & 66.42 & 75.31 & 76.62 \\
&HOS-Net & 67.51 & 68.92 & 78.62 & 77.41 \\
\rowcolor{gray!20}&PMT(baseline) & 63.84 & 60.31 & 68.02 & 70.57 \\
\rowcolor{gray!20}
&BIT (ours) & \textbf{77.51}(+13.67) & \textbf{78.32}(+18.01) & \textbf{85.71}(+17.69) & \textbf{86.43}(+15.86) \\
\hline

\end{tabular}
\label{tab1}
\end{table*}

To further validate BIT’s effectiveness toward data imbalance, we simulated controlled imbalance experiments by reducing training samples per identity on the SYSU-MM01 dataset, whose original visible/infrared ratio in the training set is 1.9. We consider four settings: randomly removing 10\% and 20\% of infrared images (which exacerbate the imbalance), and removing 10\% and 20\% of visible images (which alleviate it). Note that reducing training data inevitably causes some performance drop for all methods. The results are shown in Tab.~\ref{tab1}.

Across all four settings, BIT consistently achieves the best performance on all metrics compared with current SOTA methods. We also observe that existing rigid feature learning approaches suffer significantly larger degradation when infrared samples are reduced than when an equal proportion of visible samples is removed (\textit{e.g.}, 10\% inf reduce setting v.s. 10\% vis reduce setting), even though the absolute number of removed visible images is higher. This indicates that rigid feature learning methods are inherently sensitive to data imbalance.

In contrast, BIT remains markedly stable under all imbalance settings. Moreover, the performance gain of BIT over the baseline PMT becomes even larger in the more severely imbalanced infrared reduction settings, further demonstrating BIT’s strong robustness to uneven modality distributions and its clear advantage over prior SOTA methods. In addition, we also observe that BIT yields a larger performance gain over the baseline even when the visible samples are reduced. This indicates that BIT is not only robust to data imbalance, but also resilient to reductions in the overall training data scale.

\section{More Visualization Analysis}

\begin{figure*}[t]
\centering
\includegraphics[width=2.0\columnwidth]{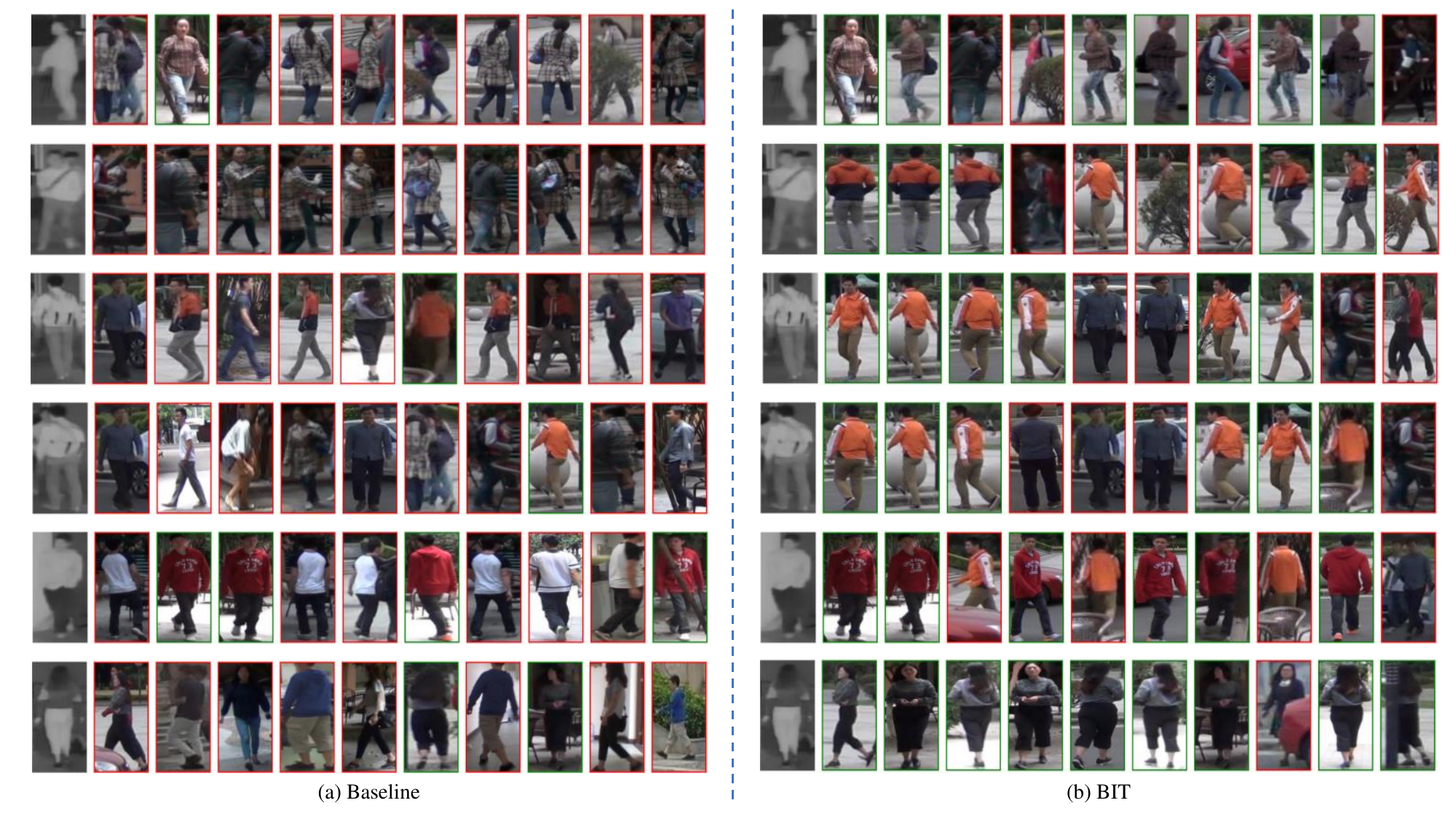}
\caption{Illustration of Rank-10 retrieval results on SYSU-
MM01. The left is baseline, the right is our BIT.}
\label{fig:supp2}
\end{figure*}
\subsection{More Retrieval Results}
To qualitatively evaluate the effectiveness of our proposed BIT framework, we present additional retrieval results on the SYSU-MM01 dataset in Fig.~\ref{fig:supp2}, comparing BIT with the baseline method. For each query image, we visualize the top-10 retrieved gallery images, where green boxes denote correctly matched identities and red boxes indicate incorrect matches.
\begin{figure}[t]
\centering
\includegraphics[width=1.0\columnwidth]{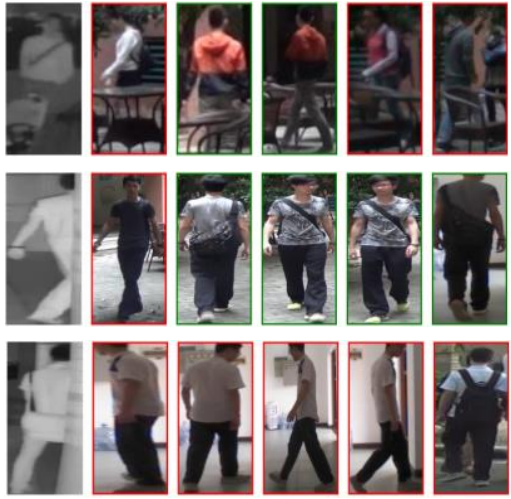}
\caption{Illustration of some unsuccessful retrieval results.}
\label{fig:supp3}
\end{figure}
As shown in the figure, BIT consistently yields more accurate retrievals, with a greater number of correct matches appearing in the top ranks compared to the baseline. This improvement highlights BIT’s enhanced ability to bridge the modality gap and capture identity-discriminative features, even under challenging cross-modality conditions. In particular, BIT demonstrates superior robustness in complex scenes where the baseline tends to misidentify hard negatives or be distracted by modality-specific cues.

Notably, in the third row, we observe that the baseline method retrieves gallery images that are visually similar to the correct match but belong to incorrect identities. This behavior indicates that the baseline relies on a rigid feature extraction paradigm, which struggles to model the complex and nonlinear relationships inherent in cross-modality scenarios, as discussed in the main manuscript. In contrast, BIT successfully avoids such false positives by directly computing the similarity between query and gallery instances in a more adaptive and identity-aware manner, retrieving the correct matches despite appearance-level distractions.

These qualitative results further corroborate the quantitative improvements reported in the main paper, demonstrating that BIT not only improves performance metrics but also enhances the visual reliability and interpretability of retrieval outcomes in practical applications.

\subsection{Some Unsuccessful Retrieval Results}
Fig.~\ref{fig:supp3} presents several representative failure cases of our BIT framework. As shown, most of these unsuccessful retrievals occur in scenarios involving severe occlusion, where critical identity cues are partially blocked. These cases reveal a compounded challenge: not only must the model bridge the modality gap between visible and infrared images, but it must also cope with significant spatial and semantic information loss caused by occlusion.

The current design of BIT primarily focuses on query aware matching. However, these results suggest that under heavy occlusion, even matching-based method is insufficient when essential visual cues are missing or distorted.

Addressing such challenging cases remains an open problem. Future work could explore integrating occlusion-aware modeling mechanisms, such as part-based feature reasoning, visibility prediction, or generative completion, which can enhance the model’s robustness under partial observation. In particular, designing solutions that can jointly reason about modality heterogeneity and occlusion patterns may offer a promising direction for advancing visible-infrared person re-identification in real-world environments.

% WARNING: do not forget to delete the supplementary pages from your submission 
% \input{sec/X_suppl}

\end{document}